\documentclass[9pt,onecolumn]{pnas-new}
\usepackage{latexsym,amssymb,amsfonts,mathrsfs,color,amsmath,wasysym,bm}
\usepackage{graphicx}
\usepackage{booktabs}
\usepackage{adjustbox}
\usepackage{lscape}
\usepackage{array} 
\usepackage{dcolumn}
\usepackage{todonotes}
\usepackage{makecell}

 \templatetype{pnasmathematics} % Choose template
\begin{document}

% \title{Locomotion Kinematics of Zoospore-Inspired Robots: Propulsion Mechanisms and Flagellar Dynamics at Low Reynolds Numbers}

% \title{Flagellar Swimming at Low Reynolds Numbers:\\
% A Zoospore-Inspired Robot Unlocks Nature's Secrets for High-Speed Propulsion}

\title{Flagellar Swimming at Low Reynolds Numbers:\\Zoospore-Inspired Robotic Swimmers with Dual Flagella for High-Speed Locomotion}

% \title{Unique Morphology and Flagellar Dynamics of Zoospores Inspire a Centimeter-Scale Biflagellated Robot Operating at Low Reynolds Numbers}

% Use letters for affiliations, numbers to show equal authorship (if applicable), and to indicate the corresponding author
\author[a]{Nnamdi C. Chikere}
\author[b]{Sofia Lozano Voticky}
\author[c]{Quang D. Tran}
\author[a, 1]{Yasemin Ozkan-Aydin}

\affil[a]{Department of Electrical Engineering, University of Notre Dame, Notre Dame, IN, US}
\affil[b]{Department of Biomedical Engineering, Pontificia Universidad Católica de Chile, Santiago, Chile}
\affil[c]{Department of Physics, Brandeis University, Waltham, MA 02453, US}

% Please give the surname of the lead author for the running footer
\leadauthor{Chikere}

% Please add a significance statement to explain the relevance of your work
\significancestatement{Zoospores utilize remarkable locomotion and survival strategies, allowing them to swim at high speeds to efficiently spread and locate new hosts while conserving their limited energy reserves. Drawing inspiration from this natural system, we introduce a centimeter-scale robotic swimmer that mimics zoospore movement using a dual-flagella system, incorporating both anterior and posterior flagella for propulsion. Through a combination of experimental and theoretical analyses, we reveal the intricate relationships between flagellum length, beating frequency, and the contributions of each flagellum, providing insights into how these elements influence overall performance. This work provides valuable insights into the hydrodynamics of biflagellated systems, opening pathways for designing advanced, efficient microscale robotics.}

% Please include corresponding author, author contribution and author declaration information
\authorcontributions{N.C.C. and S.L.V. built the robot, performed the experiments, analyzed the experimental data, Q.D.T. derived the mathematical model and analyzed the data, Y.O.A. designed the research, N.C.C., Q.D.T. and Y.O.A. wrote the paper. }
\authordeclaration{The authors declare no competing interest.}
% \equalauthors{\textsuperscript{1} }
\correspondingauthor{\textsuperscript{2}To whom correspondence should be addressed. e-mail: yozkanay@nd.edu}

% At least three keywords are required at submission. Please provide three to five keywords, separated by the pipe symbol.

\begin{abstract}

Traditional locomotion strategies become ineffective at low Reynolds numbers, where viscous forces predominate over inertial
forces. To adapt, 
microorganisms have evolved specialized structures like cilia and flagella for 
efficient maneuvering in viscous environments. Among these organisms, \textit{Phytophthora} zoospores demonstrate unique locomotion mechanisms that allow them to rapidly spread and attack new hosts while expending minimal energy. 
In this study, we present the design, fabrication, and testing of a zoospore-inspired robot, which leverages dual flexible flagella and oscillatory propulsion mechanisms to emulate the natural swimming behavior of zoospores. Our experiments and theoretical model reveal that both flagellar length and oscillation frequency strongly influence the robot's propulsion speed, with longer flagella and higher frequencies yielding enhanced performance. Additionally, the anterior flagellum, which generates a pulling force on the body, plays a dominant role in enhancing propulsion efficiency compared to the posterior flagellum's pushing force. This is a significant experimental finding, as it would be challenging to observe directly in biological zoospores, which spontaneously release the posterior flagellum when the anterior flagellum detaches. This work contributes to the development of advanced microscale robotic systems with potential applications in medical, environmental, and industrial fields. It also provides a valuable platform for studying biological zoospores and their unique locomotion strategies.
 
\end{abstract}

% \dates{This manuscript was compiled on \today}
% \doi{\url{www.pnas.org/cgi/doi/10.1073/pnas.XXXXXXXXXX}}

\maketitle
% \thispagestyle{firststyle}
% \ifthenelse{\boolean{shortarticle}}{\ifthenelse{\boolean{singlecolumn}}{\abscontentformatted}{\abscontent}}{}

% \firstpage[15]{2}
% Use \firstpage to indicate which paragraph and line will start the second page and subsequent formatting. In this example, there are a total of 11 paragraphs on the first page, counting the first level heading as a paragraph. The value {12} represents the number of the paragraph starting the second page. If a paragraph runs over onto the second page, include a bracket with the paragraph line number starting the second page, followed by the paragraph number in curly brackets, e.g. "\firstpage[4]{11}".

% If your first paragraph (i.e. with the \dropcap) contains a list environment (quote, quotation, theorem, definition, enumerate, itemize...), the line after the list may have some extra indentation. If this is the case, add \parshape=0 to the end of the list environment.
\dropcap{T}he study of microscale locomotion in fluids has garnered significant interest due to its potential applications in medical diagnostics \cite{sitti_biomedical_2015, aziz_medical_2020}, targeted drug delivery systems \cite{luo_micro-nanorobots_2018, wang_microrobots_2023}, and environmental monitoring and conservation \cite{preetam_nano_2024, beladi-mousavi_maze_2021}. At low Reynolds number environments ($\textit{Re} \ll 1$), where viscous forces dominate inertial forces \cite{purcell_life_1977}, conventional locomotion strategies employed by larger organisms become ineffective. To achieve swimming in regimes with this fluidic behavior, also known as the Stokes-flow regime \cite{stokes_mathematical_2009}, most microorganisms dwelling in this habitat have evolved either the cilia or flagella \cite{brennen_fluid_1977}, which are protruding slender, hair-like structures capable of oscillatory beating motion. They have devised different techniques for propulsion: extended groups of cilia in ciliary motion commonly found in organisms like \textit{Paramecia} and \textit{Tetrahymena}, prokaryotic flagellar motion characterized by a corkscrew rotation of the helical flagella found in bacteria like \textit{E. coli} and \textit{Salmonella typhimurium} \cite{lauga_hydrodynamics_2008}, and the whiplike or bending eukaryotic flagellar motion found in spermatozoa cells \cite{gong_steering_2020}, green algae \textit{Chlamydomonas} \cite{lewin_studies_1952} and zoospores \cite{tran_coordination_2022, judelson_spores_2005}. These specialized structures enable locomotion through non-reciprocal propulsion techniques, achieving net displacement by breaking symmetry in their motion, as described by Scallop's theorem \cite{purcell_life_1977}.
\par

Among eukaryotes, zoospores exhibit locomotion mechanisms such as forward movement and turning that enable them to navigate efficiently through water in the soil. Due to their microscopic size, they operate at low Reynolds numbers. The structure and arrangement of flagella are defining features that distinguish traits across species. For example, \textit{opisthokonts} possess one or more posterior 'whiplash' flagella \cite{money_chapter_2016}, while \textit{anisokonts} are biflagellated, featuring anterior and posterior flagella of distinct lengths arranged for differential propulsion. This study specifically examines the \textit{Phytophthora} zoospores, a class of \textit{heterokonts} \cite{yoon_stramenopiles_2009} that display coordinated flagellar motions extensively studied in previous literature \cite{walker_zoospore_2007, ochiai_pattern_2011, tran_coordination_2022}.

% The design of our zoospore-inspired robot mirrors the unique locomotion mechanisms of \textit{Phytophthora} zoospores (Fig. \ref{fig:robotParts}A), which employ a biflagellated system for efficient swimming and precise maneuvering in viscous environments. 

% Biological zoospores use two distinct flagella—a mastigoneme-decorated anterior and a smooth posterior—that facilitate rapid movement and precise navigation through tight spaces in the soil. 
\begin{figure*}[!t]
    \centering
    \includegraphics[width=0.95\textwidth]{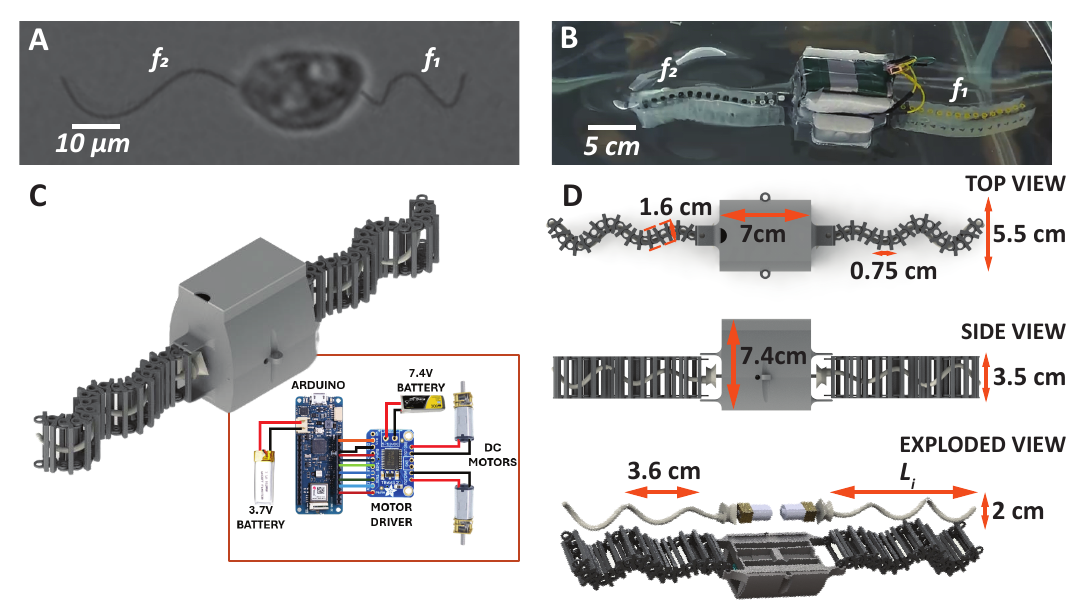}
     \caption{
     \textbf{Zoospore-inspired robot design} (A) Microscopic image of a biological zoospore showing its body and two flagella, anterior (\(f_1\)) and posterior (\(f_2\)). (B) Zoospore-inspired robot with anterior (\(f_1\)) and posterior (\(f_2\)) flagella, shown in a glycerine tank. (C) Isometric CAD view of the robot. The inset shows the electronics. (D) Dimensions of the zoospore robot: Top view, side view, and exploded view.
     }
     \label{fig:robotParts}
     % \vspace{-5 mm}
\end{figure*}
\textbf{\textit{Phytophthora} zoospores} are characterized by two distinct flagella that attach at specific points on the cell body and a kidney-like (Fig. \ref{fig:robotParts}A), ellipsoidal cell body with tapered ends, typically measuring around 10 \(\mu\)m and featuring a distinct groove along the surface \cite{tran_coordination_2022}. The anterior flagellum, approximately 15.5 \(\pm\) 0.1 \(\mu\)m in length, emerges near the groove, while the posterior flagellum, measuring about 20.3 \(\pm\) 0.76 \(\mu\)m, attaches directly to the groove—that facilitate rapid movement and precise navigation through tight spaces in the soil. The anterior flagellum has mastigonemes for thrust reversal, while the posterior flagellum is a smooth, whiplash filament. Together, these flagella's periodic beating and outward wave propagation generate high-speed forward propulsion and enable active turning. These swimmers can reach speeds of up to 100 – 400~$\mu$m/s, far surpassing the locomotion capabilities of many other microorganisms. For example, monotrichous bacteria like \textit{V. cholerae} typically reach speeds between 20 – 100~$\mu$m/s \cite{xu_role_2024}, while quadriflagellate algae such as \textit{Tetraselmis sp.} can achieve speeds of 100 – 250~$\mu$m/s \cite{yoest_barriers_2022}.

 As agents of disease spread \cite{judelson_spores_2005}, \textit{Phytophthora} zoospores are naturally optimized for high-speed, low-energy locomotion, allowing them to efficiently locate and colonize new hosts despite their limited energy reserves. Unlike other organisms that continuously produce energy, zoospores rely entirely on finite internal nutrient stores, constraining their active swimming period. By swimming faster, they can cover more ground in search of a host while conserving energy over time. This swift movement is crucial for dispersal and host colonization, as zoospores must act within a narrow window of energy availability to initiate infection successfully \cite{ochiai_pattern_2011}. The need for efficient, high-speed locomotion in zoospores has inspired our robot's design, which similarly aims to achieve fast, energy-efficient propulsion in viscous environments.

% Zoospores prioritize maximizing their chances of locating a suitable host before their limited energy reserves run out \cite{walker_zoospore_2007}.
%Notably, their locomotion strategies allow speeds of 100 – 400 \(\mu m/s\) — considerably faster than bacterial or algal movement.% 

Studying microorganisms like zoospores and their unique locomotion strategies reveals insights into life at low Reynolds numbers and inspires advancements in robotic design for high-viscosity environments. The adaptation of these biological mechanisms into robotic systems offers a promising path for enhancing robotic locomotion functionality in fluidic environments, as underscored by recent advancements \cite{palagi_bioinspired_2018, begey_manipulability_2020, li_micronanorobots_2017, sitti_biomedical_2015}. The literature on bio-inspired micro-robots is rich with examples of how biological locomotion strategies inform robotic design. For instance, studies on bacterial and algal flagella have inspired robots that replicate these microorganisms' propulsion mechanisms, achieving impressive locomotive efficiency in viscous fluids \cite{dreyfus_microscopic_2005, temel_characterization_2014, diaz_minimal_2021, chikere_effect_2023}. While bacterial and algal models are well-studied, zoospores have received relatively little attention. Additionally, scaled-up versions of flagella- and cilia-inspired robots serve as experimental models for examining locomotion mechanics and hydrodynamic behavior, providing insights into optimizing propulsion efficiency in viscous environments \cite{lim_fabrication_2022}. Regardless of size, macroscopic robots operating at low Reynolds numbers remain governed by Stokes flow, allowing them to mimic the principles of microscale propulsion effectively.

In this paper, we present the design, fabrication, and testing of a centimeter-scale robot inspired by the biflagellated locomotion of zoospores, focusing on its propulsion kinematics at low Reynolds numbers. The objectives of this study are to \textbf{(i)} understand the physical principles underlying the efficient locomotion of zoospores, particularly their use of both anterior and posterior flagella,  \textbf{(ii) }develop a theoretical model and experimentally validate the effects of key flagellar parameters — including length, amplitude, beating frequency, and the distinct contributions of each flagellum—on the robot's propulsion, \textbf{(iii) }implement these principles in the design of a bio-inspired robotic system, and \textbf{(iv)} assess the robot's overall performance in low Reynolds number fluid environments. By examining the locomotion kinematics of zoospores, we present a new approach in bio-inspired robotics that emphasizes high-speed, energy-efficient designs—features essential for advancing microscale robotic applications in viscous environments.

\section*{Results}
\subsection*{A Biflagellated Zoospore Robot}

Incorporating biological observations into our design, we developed an untethered, underactuated, low-cost, macro-sized biflagellated robot capable of traversing highly viscous fluids (Fig \ref{fig:robotParts}B.  
The mechanical design of our robotic system, as shown in Fig. \ref{fig:robotParts}C (movie S1), comprises two primary components: the rigid body housing the electronics and a pair of flagella adapted from the previous study \cite{zarrouk_single_2016}. These flagella have helical structures and are attached at both ends of the robot's body (see section Materials and Methods for details). The rotation of flagella is translated into wave-like motion through a series of interconnected links, effectively simplifying the complex natural propulsion system into a practical, bio-inspired model.

% The mechanical design of our robotic system, as shown in Fig. \ref{fig:robotParts}C, comprises two primary components: the rigid body housing the electronics and a pair of flagella positioned on the front and back of the body designed to generate propulsion through a sinusoidal wave-like movement. 

\begin{figure*}[!t]
	\centering
	\includegraphics[width=0.7\linewidth]{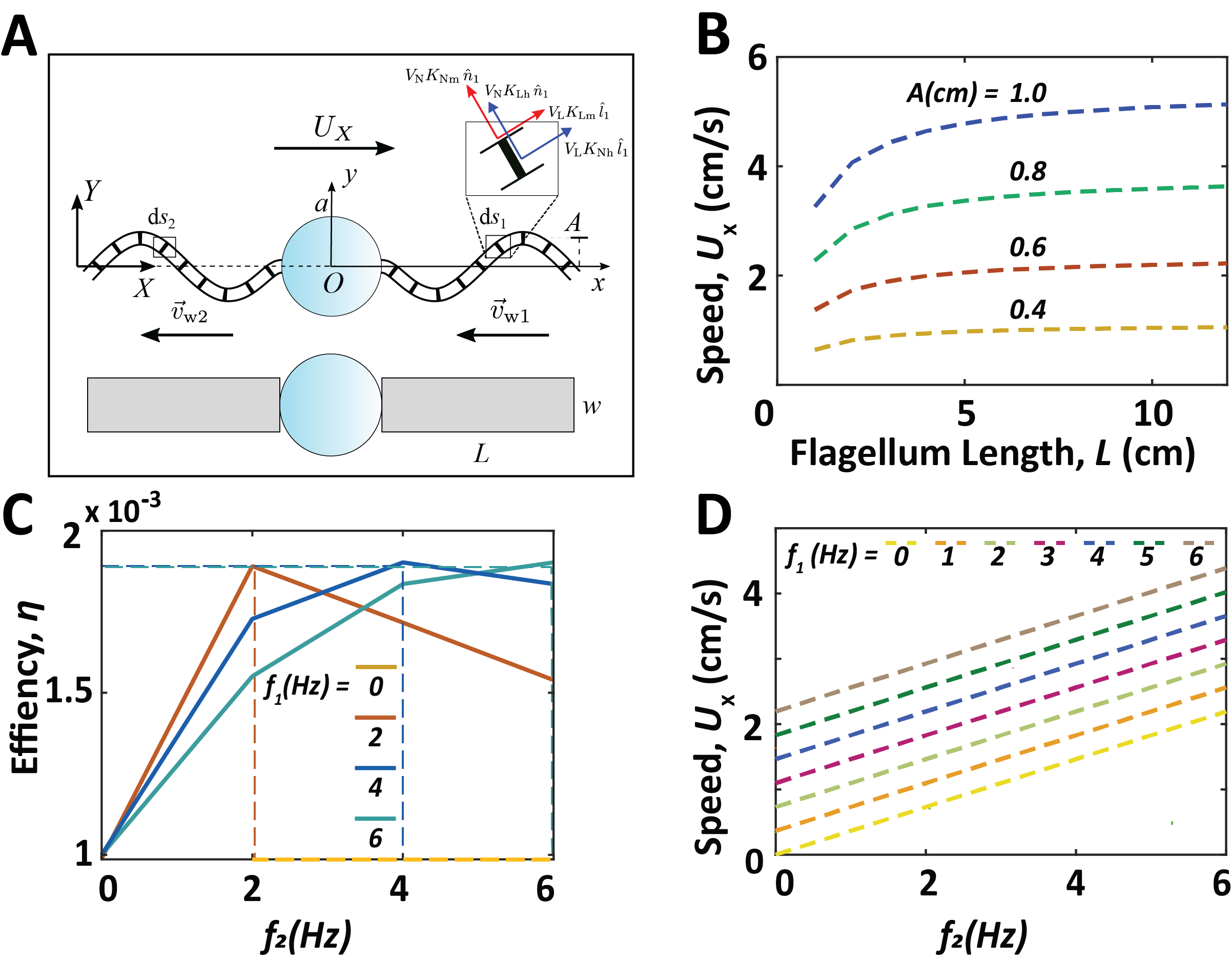}
	\caption{\textbf{Theoretical model of swimming zoospore robot using Resistive Force Theory.} (A) Schematic representation of the zoospore robot swimming in a viscous fluid, modeled with a spherical body and two flagella, beating in a sine waveform. (B) The theoretical speed of the zoospore robot as a function of flagellum length ($L$) and amplitude ($A$) for different wave amplitudes ranging from 0.4 cm to 1.0 cm. (C) Propulsion efficiency of the robot at varying anterior ($f_1$) and posterior ($f_2$) flagellum frequencies for different frequencies from 0 to 6 Hz. (D) The speed of the robot as a function of different anterior ($f_1$) and posterior ($f_2$) flagellum frequencies, ranging from 0 to 6 Hz.}
	\label{fig:model}
\end{figure*}

\begin{figure*}[!h]
    \centering
    \includegraphics[width=0.95\textwidth]{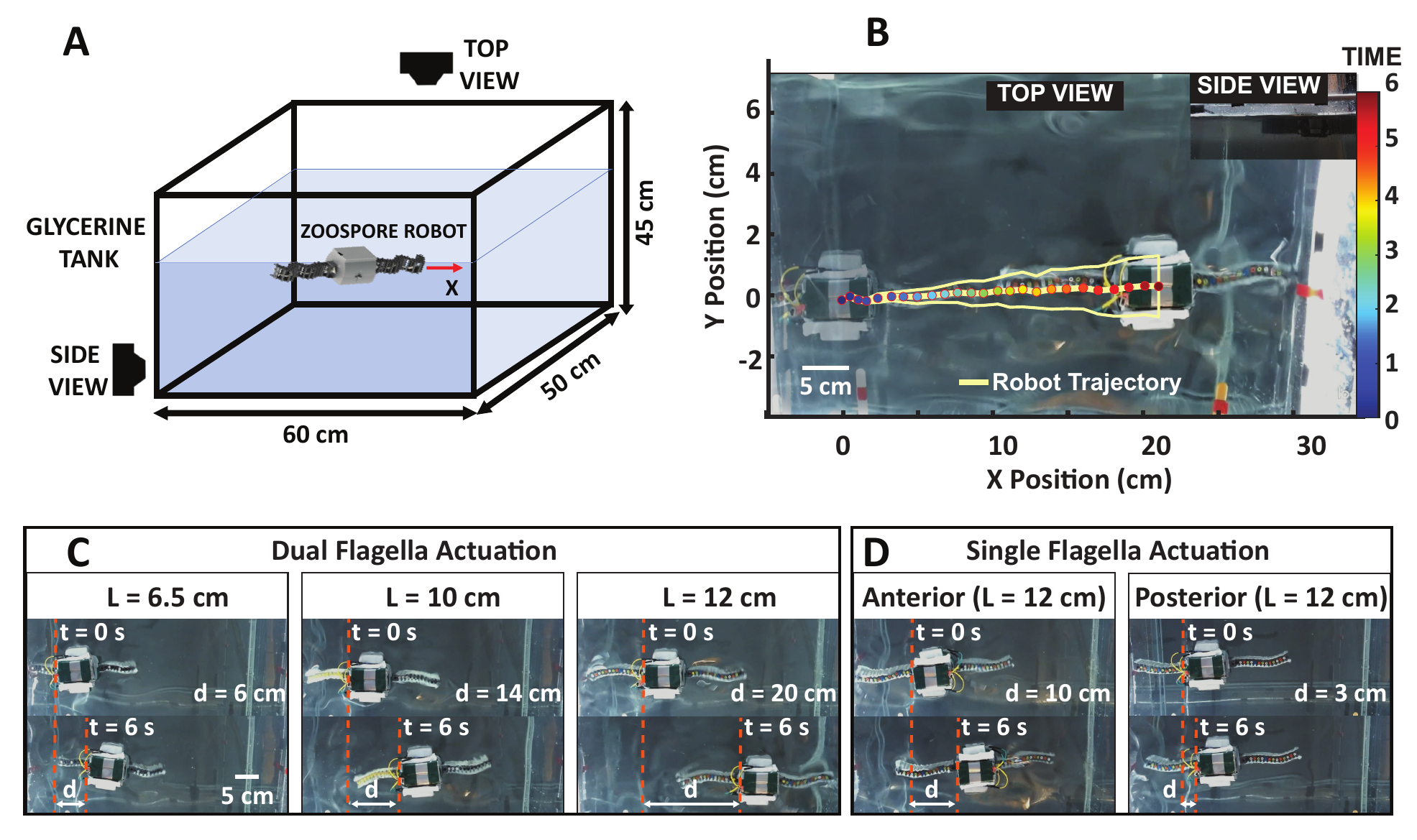}
     \caption{
     \textbf{Experimental Setup and Flagella Configurations.}(A) Schematic of the experimental setup, showing the zoospore robot in a glycerine tank from both top and side views. The robot moves along the X-axis. (B) Top view of the robot's trajectory over 6 seconds, with color coding representing time. The inset shows the side view of the robot's path over time. (C) Dual flagella actuation experiments for different flagella lengths: L = 6.5 cm, L = 10 cm, and L = 12 cm. The displacement covered in 6 seconds is shown below each configuration (6 cm, 14 cm, and 20 cm, respectively). (D) Single flagella actuation experiments for the anterior and posterior flagella with L = 12 cm. The anterior flagella achieved a 10 cm displacement in 6 seconds, while the posterior flagella covered 3 cm in the same duration.
     }
     \label{fig:expsetup}
     % \vspace{-5 mm}
\end{figure*}

\subsection*{Low Reynolds Number Fluid Interaction Modelling}
We developed a theoretical model to analyze how the beating of the two flagella drives the robot's movement (Fig. \ref{fig:model}). By modeling the fluid interaction forces, we can evaluate how varying parameters such as beat frequency, amplitude, and flagellar length contribute to the robot's propulsion. 
\par Without loss of generality, we assume the robot has a spherical body 
%described as $x^2 + y^2 = a^2$
and two opposite flagella having a sine waveform shape.
%defined as
%\begin{equation}
%	y(x,t) = A \sin \left( (-1)^{k} \omega_{k} t + (-1)^{k} \frac{2 \pi \left[ x + (-1)^{k} a\right] }{\lambda} \right) \, 
%	\label{eqn:y}
%\end{equation}
%as $(-1)^{k} x \leq -a$, where $A$ is the amplitude, $\omega_{k}$ is the angular speed, $\lambda$ is the wavelength of the anterior ($k=1$) and the posterior flagellum ($k=2$). 
Using Resistive Force Theory (RFT), 
we divide each flagellum into an infinite number of very small segments with length d$s_{k}$ (Fig. \ref{fig:model}A), at which
%and each segment is located in the body-fixed frame ($xOy$) by a position vector $\vec{r}_{k} = x_{k} \vec{i} + y_{k} \vec{j}$, where $\vec{i}$, $\vec{j}$ are the unit vectors in $x$- and $y$-direction, respectively; $x_{k}$ and $y_{k}$ satisfy the shape equation (Equation \ref{eqn:y}) for the anterior ($k=1$) and the posterior ($k=2$). 	
%RFT states that 
the drag force by a fluid acting on an infinitesimal segment d$s$ of the flagellum is proportional to the relative velocity of the fluid to the flagellum segment \cite{gray_propulsion_1955, hancock_self-propulsion_1997}, as follows
\begin{equation}
	\frac{\mathrm{d} \vec{F}}{\mathrm{d} s} = K_{\rm N} V_{\rm N} \vec{n} + K_{\rm L} V_{\rm L} \vec{l},
	\label{eqn:dF}
\end{equation}
where $V_{\rm N}$ and $V_{\rm L}$ are two components of the relative velocity of fluid in normal and tangent direction to the flagellum segment, $K_{\rm N}$ and $K_{\rm L}$ are the drag coefficients of the flagellum in normal and tangent to the flagellum segment, $\vec{n}$ and $\vec{l}$ are the unit vectors normal and tangent to the flagellum segment, respectively.

In the case of our robot, each flagellum was constructed with two parallel membranes and multiple hinges that are arranged perpendicular to the membranes (See inset of Fig. \ref{fig:model}(A)). These hinges also act as slender filaments experiencing drag from the water. This arrangement resulted in complex drag coefficients, which are the combination of the drag from the membranes and the hinges (see Supporting Information for more details).

%\begin{equation}
%	K_{\rm N} = w (K_{\rm Nm} + n h K_{\rm Lh}),
%	\label{eqn:Kntotal}
%\end{equation}
%\begin{equation}
%	K_{\rm L} = w (K_{\rm Lm} + n h K_{\rm Nh}),
%	\label{eqn:Kltotal}
%\end{equation}

%The drag coefficients $K_{\rm Nm}$, $K_{\rm Lm}$, $K_{\rm Nh}$ and $K_{\rm Lh}$ are estimated by Brennen and Winet \cite{brennen_fluid_1977}, which depends on fluid viscosity, the wavelength and diameter of a flagellum.
%We also consider that the two flagella do not generate force in $Y$-direction, but only induce thrust in $X$-direction. Hence,
%\begin{equation}
%	\mathrm{d}\vec{F} \approx \mathrm{d} F_{X} \vec{i} = (\mathrm{d}\vec{F} \cdot \vec{i}) \vec{i}.
%	\label{eqn:dFx}
%\end{equation}

%\subsubsection*{Drag on the Robot Body}
%The spherical cell body moving with velocity \( U_{X} \) also experiences a drag force from water (following Happel and Brenner \cite{happel_low_1983})
%\begin{equation}
%	\vec{F}_{d,\text{cell}} = -6 \pi \mu a U_{X} \vec{i},
%	\label{eqn:Fdcell}
%\end{equation}
%where:
%\( \mu \) is the dynamic viscosity of the fluid, \( a \) is the radius of the spherical cell body, \( U_X \) is the velocity of the cell body in the \( x \)-direction, and \( \vec{i} \) is the unit vector in the \( x \)-direction.

%As the robot moves in low Reynolds number condition, total forces equate to zero due to approximately zero inertia.
%\begin{equation}
%	\Sigma\vec{F} = \vec{F}_{1} + \vec{F}_{2} + \vec{F}_{d,cell} = \vec{0}.
%	\label{eqn:Ftotal}
%\end{equation}
We then derive translational speed $U_{X}$ of the zoospore robot as shown in Equation \ref{eqn:Ux}
\begin{equation}
	U_{X} = \frac{-\pi^{2} \beta^{2} K_{\rm N} L (\gamma - 1)(v_{\rm w1} + v_{\rm w2})}{K_{\rm N} L (\gamma + 2 \pi^2 \beta^{2}) + 3 \pi \mu a (1 + 2 \pi^2 \beta^{2})},
	\label{eqn:Ux}
\end{equation}	
where $\mu$ is the dynamic viscosity of the fluid, $a$ is the radius of the spherical cell body, $A$ is the amplitude, $\omega_{k}$ is the angular speed, $\lambda$ is the wavelength of the flagellum, $L$ is the flagellum length, $v_{\rm{w}1}$ and $v_{\rm{w}2}$ are the wave propagation velocity of the anterior and posterior flagellum, $\gamma = K_{\rm{L}}/K_{\rm{N}}$ is the drag coefficient ratio, $\beta = A/\lambda$ is the flagellar shape coefficient.

\begin{figure*}[!h]
    \centering
    \includegraphics[width=0.8\textwidth]{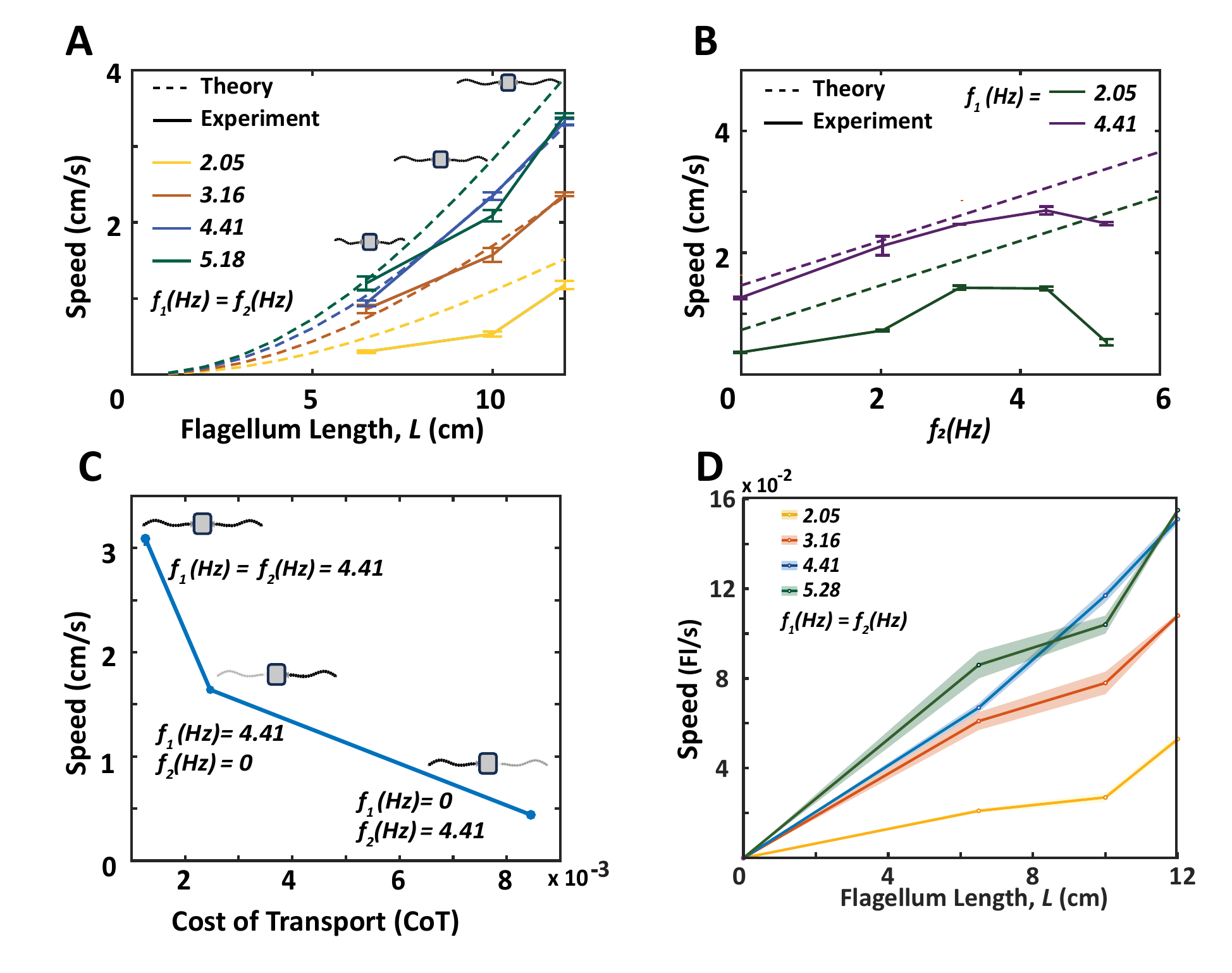}
     \caption{\textbf{Experimental and theoretical Results} (A) Theoretical (dashed lines) and experimental (solid lines) speeds of the robot as a function of flagellum length ($L$) for different actuation frequencies ($f_1 = f_2$) at 2.05, 3.16, 4.41, and 5.18 Hz. (B) Comparison of speed as a function of anterior ($f_1$) and posterior ($f_2$) flagellum for frequencies of 2.05 Hz and 4.41 Hz. (C) The speed of the robot versus Cost of Transport (CoT) for different flagella combinations - dual flagella, anterior flagella, and posterior flagella (black-actuated, and grey-turned off) (D) Speed as a function of flagellum length ($L$) normalized to body lengths per second (BL/s) for different frequencies. Error bars indicate standard deviations.}
     \label{fig:expresults}
     % \vspace{-5 mm}
\end{figure*}

% \subsection*{Dependence of propulsion velocity on flagellar beating shape}
% The model predicts that the flagellar waveform beating shape, including flagellar length, amplitude, and wavelength, can influence the robot's speed.
% The computed speed with varied flagellar length and amplitude showed that larger amplitude significantly elevated the robot speed, while increased flagellar length contributed mildly to the speed (Fig. \ref{fig:model}B). 

% However, due to mechanical fabrication constraints, each flagellum requires a certain length before fully achieving the desired amplitude. For instance, our 6.5-cm flagellum has an average amplitude of 0.4~cm, 10-cm flagellum has an average amplitude of 0.6~cm and 12-cm flagellum has an average amplitude of 0.75~cm. Zoospore robots equipped with the set of flagellar length of 12~cm obtained 3.5-times increase in speed, compared with the robots with 6.5-cm flagellar length, which was reflected accurately by our theoretical model (Fig. \ref{fig:model}(E)) Our experimental results as shown in Fig. \ref{fig:expresults}A displayed this trend as increasing the flagella length, increased the propulsion velocity, at 4.41 Hz beating frequency, the 12cm configuration recorded a speed of 3.32 $pm$ 0.04 cm/s followed by the 10cm config which has a speed of 2.35 $pm$ 0.05cm/s, last was the 6.5cm config which had a speed of 0.937 $pm$ 0.03 cm/s. 

\subsection*{Dependence of Propulsion Speed on Flagellar Beating Shape}

The model predicts that the flagellar beating shape—specifically, the length, amplitude, and wavelength—strongly influences the robot's propulsion speed. Simulations with varied flagellar length and amplitude reveal that a larger amplitude significantly increases speed, while an increase in flagellar length contributes only slightly (Fig. \ref{fig:model}B).

However, in the physical system, mechanical fabrication constraints impose a minimum flagellum length needed to achieve the desired amplitude. For instance, our 6.5~cm flagellum achieves an average amplitude of 0.4~cm, while the 10~cm and 12~cm flagella reach amplitudes of 0.6~cm and 0.75~cm, respectively. Experimental results (Fig. \ref{fig:expresults}A, movie S2) confirm this trend: at a beating frequency of 4.41~Hz, the 12~cm configuration achieved a speed of $3.32 \pm 0.04$~cm/s, followed by the 10~cm configuration at $2.35 \pm 0.05$~cm/s, and the 6.5~cm configuration at $0.94 \pm 0.03$~cm/s (Fig. \ref{fig:expsetup}C). The zoospore robot equipped with 12~cm flagella exhibited a 3.5-fold increase in speed compared to those with 6.5~cm flagella, aligning well with our theoretical predictions.

% \subsection*{Influence of Flagella Beating Frequency}
% Using our theoretical model, we calculated the robot's speed across different operating frequencies of the two flagella (Fig. \ref{fig:model}D). The model's predictions closely matched the experimental data as shown in Fig. \ref{fig:expresults}B, confirming the model's accuracy and robustness. This agreement between theory and experiment provides confidence in the model's validity, allowing us to predict and fine-tune the robot's operating conditions to achieve the desired propulsion speed. For the experiments, we considered the long 12cm flagella configuration and varied the frequency of the flagellum from 2.05 Hz TO 5.18 Hz and the experimental results (Fig. \ref{fig:expresults}A), we observed that the robot's speed increased consistently with the frequency of the flagella from 1.18 $pm$ 0.05 cm/s at 2.05Hz to 3.41 $pm$ 0.03 cm/s at 5.28Hz. We observed this increase to be consistent up to 4.41Hz, but in some experimental configurations, at a frequency of 5.18~Hz, a slight decrease in speed occurred. This decline in speed could be attributed to the motor reaching its maximum power output, which may limit the torque available at higher rotational speeds.

\subsection*{Influence of Flagella Beating Frequency}

Using our theoretical model, we calculated the robot's speed across different operating frequencies of the two flagella (Fig. \ref{fig:model}D). The model's predictions closely matched the experimental data, as shown in Fig. \ref{fig:expresults}A (movie S3), confirming its accuracy and robustness and supporting its validity for predicting and optimizing the robot's operating conditions to achieve the desired propulsion speed.

For the beating frequency experiments, we used the long 12~cm flagella configuration and varied the beating frequency of both flagella from 2.05~Hz to 5.18~Hz. Experimental results (Fig. \ref{fig:expresults}A) show that the robot's speed increased consistently with the frequency of the flagella, ranging from $1.18 \pm 0.05$~cm/s at 2.05~Hz to $3.41 \pm 0.03$~cm/s at 5.28~Hz. This increase was consistent up to 4.41~Hz, but in some experimental configurations, a slight decrease in speed was observed at 5.18~Hz. This decline could be attributed to the motor reaching its maximum power output, which may limit the torque available at higher rotational speeds.

\subsection*{Dominant Role of Anterior Flagellum in Propulsion Speed Enhancement}
The anterior flagellum plays a critical role in enhancing the propulsion speed of the zoospore robot. 
% Similarly, in biological zoospores, the anterior flagellum generates significant thrust to pull the cell forward, aided by its mastigonemes for effective propulsion, while the posterior flagellum contributes to steering and stability \cite{tran_coordination_2022}. 
%As demonstrated in our numerical model (Fig. \ref{fig:model}D), there is a consistent and significant increase in propulsion speed with rising anterior flagellar frequency. This trend was also observed in our experiments (Fig. \ref{fig:expresults}B) as when the posterior flagellum frequency is held constant at 4.41 Hz, an increase in the anterior frequency from 0 to 4.41 Hz results in a substantial rise in average speed, from $0.44 \pm 0.02$ cm/s at 0 Hz to $3.09 \pm 0.06$ cm/s at 4.41 Hz. This demonstrates the dominant role of the anterior flagellum in propulsion. However, a further increase in anterior frequency to 5.28 Hz leads to a slight decrease in speed to $2.97 \pm 0.06$ cm/s, indicating an optimal frequency range for maximum propulsion efficiency given the mechanical constraints of our robot. Hence, we kept the frequency at 4.41 Hz for our experiments as it provides the best balance between propulsion speed and energy efficiency without overloading the motors.
Our experimental results show that when the anterior frequency was held constant at 4.41 Hz, even with the posterior flagellum turned off (0 Hz), the robot still produced a significant propulsion speed of $1.64 \pm 0.02$ cm/s achieving a displacement of about 10 cm in 6 seconds (Fig. \ref{fig:expsetup}D). 
This speed significantly decreased when the posterior flagellum was propelling alone, producing an initial speed of just $0.44 \pm 0.02$ cm/s when the anterior frequency was set to 0 Hz (Table \ref{tab:power}). As the posterior frequency varies from 0 to 5.28 Hz, the increase in propulsion speed is less pronounced, with the speed improving from $1.64 \pm 0.02$ cm/s to $3.09 \pm 0.06$ cm/s as shown in Fig. \ref{fig:expresults}B (movie S4). Since the two flagella of the robot are identical, this finding indicates that the arrangement of force transmission affects the efficiency of the flagella in propelling the body through a viscous medium. In particular, pulling force by the anterior flagellum was more effective than pushing force by the posterior flagellum. When pulling an object through a viscous fluid, the force is directly applied to the object, allowing for a more efficient energy transfer. This minimizes the resistance encountered compared to pushing, where the force has to be transmitted through the fluid first, leading to greater energy dissipation.
Moreover, pulling tends to create a streamlined flow over the object, reducing the drag force and providing better stability. On the other hand, pushing may create vibration and increased drag, thus making it more challenging to move the object efficiently. These results are not obtainable in our theoretical model as we assumed that the robot only traveled in one direction and movement in other directions was neglected.
%These results underscore the dominant role of the anterior flagellum in driving propulsion, as its influence on speed is far more significant than that of the posterior flagellum. 
This observation contributes to a better understanding of the roles of each flagellum in biological zoospores \cite{tran_coordination_2022}. Our work provides experimental evidence for the biological observation that the anterior flagellum acts as a primary motor for achieving high locomotion speeds, while the posterior flagellum only contributes mildly to the swimming speed. This finding on the force contribution of each flagellum has not been experimentally verified in biological zoospores, as physical stimuli in attempts to constrain them may signal zoospores to lose both of their flagella to perform encystment \cite{andrews_asexual_1997, bassani_phytophthora_2020}.

% These results suggest that the anterior flagellum has a more direct and dominant effect on propulsion speed. Its ability to generate strong fluid displacement and forward thrust, particularly at optimal frequencies, leads to more efficient locomotion. Meanwhile, the posterior flagellum plays a secondary role, primarily contributing to stability and fine-tuning of movement rather than serving as the primary driver of propulsion. However, optimizing the coordination of both flagella results in greater overall speed than when either flagellum is used alone, highlighting the advantage of biflagellated microrobots inspired by zoospores. T

\subsection*{Dual Flagella Configuration Results in Improved Robot Speed and Efficiency}

We calculated the propulsive efficiency $\eta$ of the robot as the ratio of the power required to move the cell body at speed $U_{X}$ to the total power produced by both flagella,
\begin{equation}
    \eta = \frac{P_{\text{useful}}}{P_{\text{total}}} = \frac{F_{d,\text{cell}} \cdot U_X}{P_1 + P_2},
\end{equation}
where $F_{d,\text{cell}}$ is drag force of fluid acting on the body, $U_X$ is the robot speed in forward direction, $P_1$ and $P_2$ are the power produced by the anterior and posterior flagellum, respectively (see Supporting Information for detailed derivations). Our results indicate that both efficiency and speed are maximized when the two flagella beat at the same frequency (Fig. \ref{fig:model}C), suggesting that optimal performance is achieved when both motors operate at synchronized speeds.

By measuring each motor's power consumption during experiments, we observed that two motors operating at 4.41~Hz produced a robot speed of 3.09~cm/s, with a combined power consumption of approximately 10~W. In contrast, using a single motor at the anterior flagellum at 4.41~Hz achieved a speed of 1.64~cm/s while consuming the same amount of power (Table \ref{tab:power}). These results highlight the advantages of the dual-flagella configuration, including higher speed and improved efficiency compared to a single-flagellum setup.
\begin{table}
    \centering
    \begin{tabular}{cccc}
        & Frequency (Hz) & Speed (cm/s) & Power (W) \\
        Both flagella & 4.41 & 3.09 & 9.82\\
        Anterior flagellum & 4.41 & 1.64 & 10.22\\
        Posterior flagellum & 4.41 & 0.44 & 9.43\\
    \end{tabular}
    \caption{Speed and power consumption of the robot operating at different conditions.}
    \label{tab:power}
    \vspace{-5mm}
\end{table}
We also quantified the robot's Cost of Transport (CoT) (Fig. \ref{fig:expresults}C) across three configurations: anterior-only, posterior-only, and dual-flagella. The CoT, defined as

\begin{equation}
    \text{CoT} = \frac{P}{mgU},
\end{equation}
where \( P \) is the total power consumption, \( m \) is the mass, \( g \) is the gravitational constant, and \( U \) is the speed of the robot.
The CoT was lowest in the dual-flagella configuration, recording a value of \( 1.3 \times 10^{-3} \). In comparison, the anterior-only and posterior-only configurations resulted in higher CoTs of \( 2.5 \times 10^{-3} \) and \( 8.5 \times 10^{-3} \), respectively. These results emphasize the energy efficiency of the dual-flagella configuration, achieving optimal propulsion at a lower energetic cost, highlighting the advantage of biflagellated microrobots inspired by zoospores.

\section*{Discussion}
This work translates insights from the locomotion of \textit{Phytophthora} zoospores into the design of a dual-flagellated robotic swimmer optimized for low Reynolds number environments, enhancing speed, energy efficiency, and maneuverability while offering a versatile platform for applications in biological, robotic, and fluid dynamics research.

\begin{figure}[!t]
    \centering
  \includegraphics[width=0.7\columnwidth]{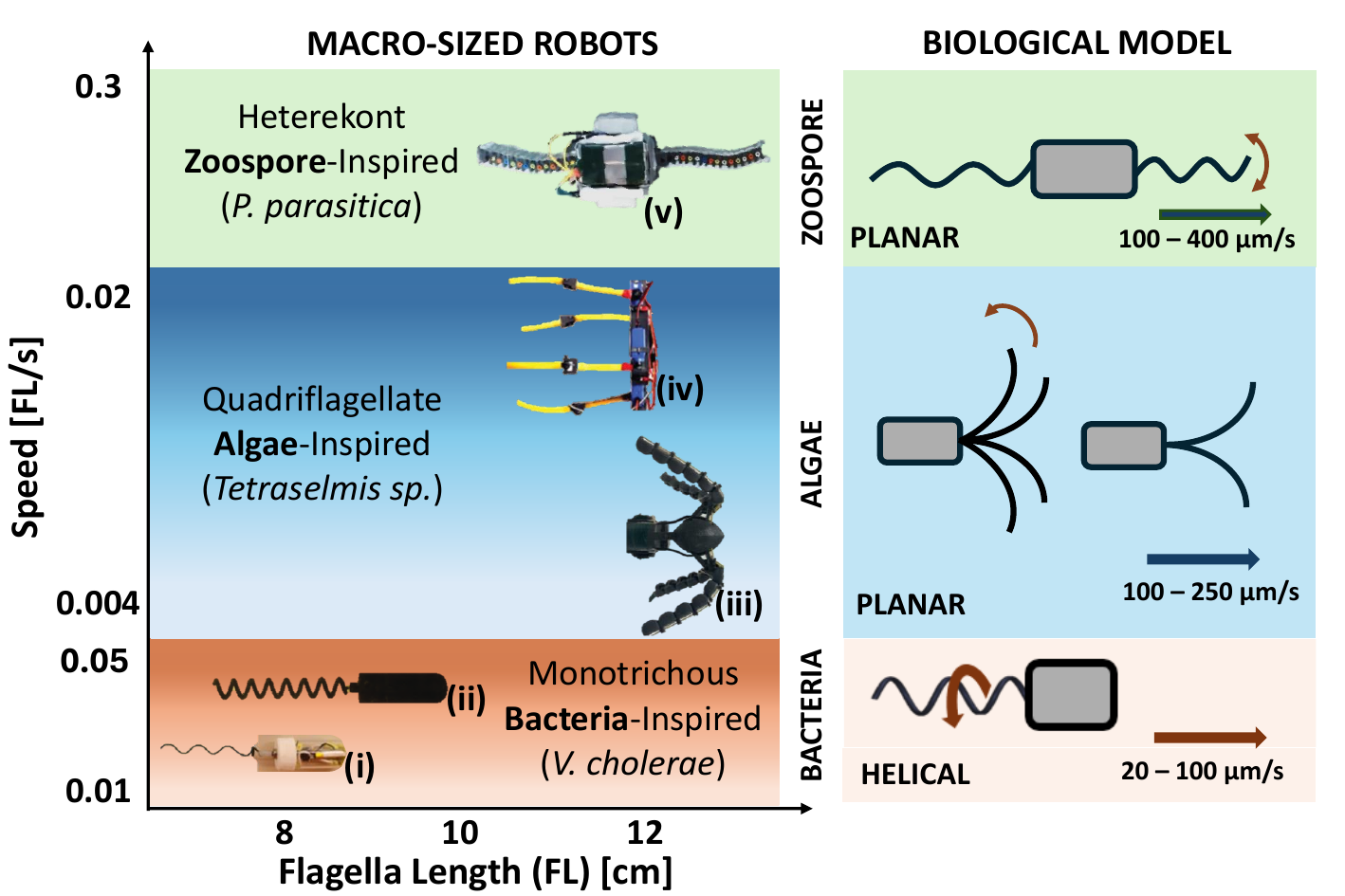}
    \caption{\textbf{Comparison of macro-sized robots and biological models based on flagella-inspired propulsion systems.} The left panel shows macro-sized robots, including (i) \cite{temel_characterization_2014} and (ii) \cite{das_force_2022} Monotrichous bacteria-inspired robots, (iii)  \cite{chikere_harnessing_2024} and (iv) \cite{diaz_minimal_2021} quadriflagellate algae-inspired robots, and (v) our zoospore-inspired robot, with their corresponding flagella lengths and speeds normalized to flagellar lengths (FL/s). The right panel displays the biological models for each category: Monotrichous bacteria with helical propulsion, quadriflagellate algae with planar propulsion, and heterokont zoospores with planar propulsion.}

    \label{fig:macrorobots}
    % \vspace{-5 mm}
\end{figure}

In comparing the forward propulsion performance of our zoospore-inspired robot to other macro-size flagellated robotic swimmers, it is evident that zoospore-inspired designs offer unique advantages in both speed and locomotion efficiency. As shown in Fig. \ref{fig:macrorobots}, flagellated robots inspired by different microorganisms exhibit varying performance levels depending on their morphology and propulsion mechanisms. These bio-inspired robotic swimmers emulate the natural speed capabilities of their biological counterparts, with the zoospore-inspired robot designed to mimic the dual-flagellated locomotion of \textit{P. parasitica}, achieving a maximum speed of 3.4~cm/s (0.3 Flagella Lengths/s). This performance surpasses that of bacteria-inspired (0.38~cm/s [0.05 Flagella Lengths/s]) and algae-inspired (0.14~cm/s [0.01 Flagella Lengths/s]) robotic designs in the same size scale.

The zoospore robot's superior speed can be attributed not only to its biflagellated design, which efficiently combines the thrust generated by both anterior and posterior flagella but also to some unique design and functional features. The robot utilizes a sheet-like structure for its flagella, which may enhance thrust generation compared to the slender rod structures used in other biflagellated robots. Additionally, the specific tuning of flagellar frequency and amplitude maximizes propulsion efficiency within the constraints of the robot's materials and motor capabilities. 
The dual-flagella configuration offers advantages beyond speed, providing enhanced locomotion capabilities and energy efficiency compared to single-flagellum designs commonly used in microscale robotics. This work builds upon foundational research in bio-inspired robotics by showing that emulating biflagellated propulsion, as observed in zoospores, can yield robotic systems that replicate the capabilities of natural swimmers more effectively in viscous environments. These findings suggest that dual-flagella designs may offer a superior solution for microscale robotics where energy conservation and directional control are critical.
Our zoospore robot enables controlled experimentation to examine how flagellar parameters—such as length, amplitude, and frequency—influence propulsion. 

Furthermore, this platform facilitates the exploration of hypotheses that are challenging to investigate biologically, such as the energetic advantages of dual flagella for swimming, a characteristic observed in zoospores but difficult to study directly due to their small size and complex fluid dynamics. By addressing gaps in our understanding of fluid dynamics at small scales, this robotic model offers valuable insights that can be scaled to microsize applications. Its simplicity and adaptability make it suitable for tasks such as inspection and exploration in confined or hazardous environments, offering a new solution for scenarios where traditional robots may struggle.

While our findings are promising, the robot's design has morphological limitations, particularly in lacking the agile turning mechanisms seen in natural zoospores, which future work could address to improve maneuverability in constrained environments. Additionally, our design employs a sheet structure for the flagella, while natural zoospores use a slender rod-like structure, which may impact propulsion dynamics. The flagellar amplitude in this study was fixed, limiting the exploration of varied amplitudes that could affect speed and efficiency. We also did not explore alternative materials for the flagella, which could offer different mechanical properties and performance outcomes. Further scaling down the robot could enable microscale medical procedures and environmental monitoring applications, where smaller and more agile robots are essential. Addressing these areas could bring the design closer to biological systems' adaptive and flexible capabilities.

In summary, the zoospore-inspired robotic platform exemplifies how biological insights can inform engineering solutions, offering a powerful tool for studying microscale propulsion and enabling new scientific investigations. Its application potential extends beyond laboratory studies, providing a foundation for microrobots capable of navigating constrained, toxic, or otherwise inaccessible environments. By bridging biology and engineering, this work contributes both to the fundamental understanding of microscale locomotion and to the practical advancement of bio-inspired robotics, setting the stage for future developments that will expand the utility and capability of microscale robotic systems.
% \vspace{-5mm}

\section*{Materials and Methods}
\label{section_MM}
\subsection*{Mechanical Design and Electronics}
The robotic body is designed in a hexagonal cylindrical form to optimize space for fitting electronic components and motors while also providing flat surfaces for precise flagella placement. The body measures 7 cm in length, 7.4 cm in height, 5.5 cm in width, and weighs 256 grams. It is segmented into two parts joined by screws. The top section houses the electronics, sealed to prevent glycerin from seeping in, while the bottom compartment holds two DC motors and is fully immersed in the fluid during operation. These motors are positioned at the same horizontal level but oriented toward opposite ends of the body, centered on each side to allow for balanced propulsion. To achieve neutral buoyancy, foam blocks are added to the sides of the robot, helping it maintain a stable position in the fluid.

The propulsion mechanism for our robot's flagella, shown in Fig. \ref{fig:robotParts}D, draws inspiration from previous work \cite{zarrouk_single_2016}. This design features an internal helix that navigates through a chain of links connected by rotational joints. The 2D projection of the helix's curve creates a sinusoidal wave pattern, with the joint structure limiting horizontal movement. The motor rotates the helix and generates an advancing wave along the flagellum.

The internal helix length varies by configuration, with the longest measuring 12 cm. All helix configurations have a 2 cm diameter and a 3.6 cm pitch, providing the primary propulsion force. Each link measures 3.5 cm in width and 1.6 cm in height, connected by rotational joints spaced 0.75 cm apart. The total number of links in each flagellum depends on the length of the helix.

To maximize the effective area of the flagella and, consequently, the force generated by them, a flexible and thin resistance band was applied to the bottom and top surfaces of each chain of links (Fig. \ref{fig:robotParts}B). All the mechanical components of the robot's body were designed using SOLIDWORKS software and subsequently fabricated using Acrylonitrile Butadiene Styrene (ABS) material with a Stratasys F170 printer.

The flagella are actuated by two DC gear motors (Polulu, 6 Volts, 100 RPM, 70 mA), selected for their balance between torque and propulsion speed. Control over the motors' speed and direction is achieved through an Arduino MKR WiFi 1010 module interfaced with an Adafruit TB6612 1.2A DC/Stepper Motor Driver. This setup was chosen to facilitate Over-The-Air (OTA) uploading of gait codes, enabling the robot to operate autonomously without needing physical connections to modify its programming. 

A 300mAh 7.4V LiPo battery powers the motors, providing a consistent power supply throughout experiments. A separate 3.7V 250mAh LiPo battery powers the Arduino, ensuring the control system remains operational even in extended testing scenarios.

\subsection*{Experimental Setup}

To replicate the microscale environment typical for zoospore movement, vegetable glycerine (Bulk Apothecary\copyright) was chosen for its high viscosity of 1.49 $\rm Pa\cdot s$, approximately 1500 times that of water, coupled with a density of 1000 $kg/m^3$ at 20$^\circ$C. This selection aimed to simulate the highly viscous habitats in which these microorganisms thrive. The setup, as depicted in Fig. \ref{fig:expsetup}A, consisted of a transparent acrylic tank (dimensions: 60x50x45 cm$^3$) sourced from CLEAR 2420 by Waterbox Aquariums\copyright, filled to two-thirds capacity with the chosen glycerine solution.

To observe and record the robot's locomotive behavior, two high-definition cameras (Logitech C920x HD Pro) were secured above and on the side of the tank to capture top-view and side-view videos of the robot's movements at a resolution of 30 frames per second. This setup provided a comprehensive view of the robot’s propulsion and navigation through the viscous glycerine medium. During each trial, the robot was activated and placed at one end of the glycerine-filled tank, and its progress was recorded. Each experimental condition—such as varying flagellar lengths, beating frequencies, and flagellar arrangements—was repeated across three trials to ensure consistency and allow for statistical averaging.

The recorded videos were analyzed using custom MATLAB scripts along with the Tracker video analysis and modeling tool, which tracked the robot's position over time to calculate its speed, displacement, and trajectory in the \(x\)-plane (Fig. \ref{fig:expsetup}B). Key metrics, such as propulsion speed and path linearity, were determined by tracking the center of mass of the robot in each frame. For each condition, the average speed was calculated along with standard deviations to evaluate consistency across trials. After data collection, the impact of each variable (e.g., flagellar frequency, length) on propulsion efficiency was analyzed and visualized through plots, providing insights into the conditions that produced optimal propulsion performance.

\showmatmethods{} % Display the Materials and Methods section

\acknow{S.L.V. was supported by the Bridge Program  of the Pontifical Catholic University (PUC) and the University of Notre Dame.}

\showacknow{} % Display the acknowledgments section

% \bibsplit[2]
%Use \bibsplit to split the references from the body of the text. Value "[2]" represents the number of reference in the left column (Note: Please avoid single column figures & tables on this page.)

% Bibliography

% \bibliography{references}

\end{document}

\maketitle

%% Comment out or remove this line before generating final copy for submission; this will also remove the warning re: "Consecutive odd pages found".
% \instructionspage  

%% Adds the main heading for the SI text. Comment out this line if you do not have any supporting information text.
\SItext

\section*{Low Reynolds number fluid interaction modeling}
We develop a theoretical model to study how the beating of the two flagella results in the movement of the robot (Fig. 2A, main text).
Without losing generality, we assume the robot has a spherical body and two opposite flagella having sine waveform shape defined as
\begin{equation}
	y(x,t) = A \sin \left( (-1)^{k} \omega_{k} t + (-1)^{k} \frac{2 \pi \left[ x + (-1)^{k} a\right] }{\lambda} \right) \, 
	\label{eqn:y}
\end{equation}
as $(-1)^{k} x \leq -a$, where $A$ is the amplitude, $\omega_{k}$ is the angular speed, $\lambda$ is the wavelength of the anterior ($k=1$) and the posterior flagellum ($k=2$). 

Following this method, each flagellum is divided into an infinite numbers of very small segments with length d$s_{k}$, and each segment is located in the body-fixed frame ($xOy$) by a position vector $\vec{r}_{k} = x_{k} \vec{i} + y_{k} \vec{j}$, where $\vec{i}$, $\vec{j}$ are the unit vectors in $x$- and $y$-direction, respectively; $x_{k}$ and $y_{k}$ satisfy the shape equation (Equation \ref{eqn:y}) for the anterior ($k=1$) and the posterior ($k=2$).
	
RFT states that the drag force by fluid acting on an infinitesimal segment d$s$ of the flagellum is proportional to the relative velocity of fluid to the flagellum segment \cite{gray1955propulsion,hancock1953self,koh2016theoretical}, as follows
\begin{equation}
	\frac{\mathrm{d} \vec{F}}{\mathrm{d} s} = K_{\rm N} V_{\rm N} \vec{n} + K_{\rm L} V_{\rm L} \vec{l},
	\label{eqn:dF}
\end{equation}
where $V_{\rm N}$ and $V_{\rm L}$ are two components of relative velocity of fluid in normal and tangent direction to the flagellum segment, $K_{\rm N}$ and $K_{\rm L}$ are the drag coefficients of the flagellum in normal and tangent to the flagellum segment, $\vec{n}$ and $\vec{l}$ are the unit vectors normal and tangent to the flagellum segment, respectively.

In case of our robot, each flagellum was constructed with two membranes in parallel and multiple lateral hinges. Thus, each segment of the flagellum possesses a complex drag coefficient as derived below

\begin{equation}
	K_{\rm N} = w (K_{\rm Nm} + n h K_{\rm Lh}),
	\label{eqn:Kntotal}
\end{equation}
\begin{equation}
	K_{\rm L} = w (K_{\rm Lm} + n h K_{\rm Nh}),
	\label{eqn:Kltotal}
\end{equation}
where $K_{\rm Nm}$, $K_{\rm Lm}$ are the drag coefficients of the membrane in normal and tangent to the flagellum segment, $K_{\rm Nh}$, $K_{\rm Lh}$ are the drag coefficients of the hinges in normal and tangent to the flagellum segment, $w$ is the width of the membrane, $h=1.6$~cm is the length of the hinge, $n \approx 2$ is the density of the hinge in 1~cm flagellum length.

The drag coefficients $K_{\rm Nm}$, $K_{\rm Lm}$, $K_{\rm Nh}$ and $K_{\rm Lh}$ are estimated by Brennen and Winet \cite{brennen1977fluid}, which depends on fluid viscosity, the wavelength and diameter of a flagellum.

\begin{equation}
	K_{\rm N} = \frac{4 \pi \mu}{\ln\left( \frac{4 \lambda}{d}\right) -2.90},
	\label{eqn:Kn}
\end{equation}
\begin{equation}
	K_{\rm L} = \frac{2 \pi \mu}{\ln\left( \frac{4 \lambda}{d}\right) -1.90},
	\label{eqn:Kl}
\end{equation}
where $\mu = 1.49$~Pa.s is the viscosity of glycerin, $\lambda$ is the wavelength of the flagellum, $d$ is the diameter of the flagellum.

We also consider that the two flagella do not generate force in $Y$-direction, but only induce thrust in $X$-direction. Hence,
\begin{equation}
	\mathrm{d}\vec{F} \approx \mathrm{d} F_{X} \vec{i} = (\mathrm{d}\vec{F} \cdot \vec{i}) \vec{i}.
	\label{eqn:dFx}
\end{equation}

Following the mathematical model in the previous work of one of us \cite{tran2022}, we derive the drag force of fluid acting on each flagellum as

\begin{equation}
	F_{X,1} = K_{\rm N} L \left[ \frac{ -2 \pi^{2} v_{\rm w1} (\gamma -1) \beta^{2} - (\gamma -1) U_{X}}{1 + 2 \pi^2 \beta^{2}} - U_{X} \right],
	\label{eqn:Fx1}
\end{equation}

and

\begin{equation}
	F_{X,2} = K_{\rm N} L \left[ \frac{-2 \pi^{2} v_{\rm w2} (\gamma -1) \beta^{2} - (\gamma -1) U_{X}}{1 + 2 \pi^2 \beta^{2}} - U_{X} \right],
	\label{eqn:Fx2}
\end{equation}

where $v_{\rm w1} = \lambda f_{1}$ and $v_{\rm w2} = \lambda f_{2}$ are the wave propagation velocity of the anterior and posterior flagellum, respectively, $\gamma = K_{\rm L}/K_{\rm N}$, $\beta = A/\lambda$.

At the same time, the ellipsoidal cell body moving with velocity $U_{X}$ also experiences a drag force from fluid (following Happel and Brenner)
\begin{equation}
	\vec{F}_{d,cell} = -6 \pi \mu b U_{X} \vec{i},
	\label{eqn:Fdcell}
\end{equation}

In low Reynolds number condition, total forces equate to zero due to approximately zero inertia.

\begin{equation}
	\Sigma\vec{F} = \vec{F}_{1} + \vec{F}_{2} + \vec{F}_{d,cell} = \vec{0}.
	\label{eqn:Ftotal}
\end{equation}

We derive translational speed $U_{X}$ of the zoospore as shown in Equation \ref{eqn:Ux}
\begin{equation}
	%U_{X} = \frac{ -\frac{2 \pi^{2} \beta^{2} K_{\rm N} L (\gamma - 1)  }{1 + 2 \pi^2 \beta^{2} } (v_{\rm w1} + v_{\rm w2})}{ 2 K_{\rm N} L \left( \frac{\gamma - 1}{1 + 2 \pi^2 \beta^{2}} +1 \right) + 6 \pi \mu a},
    U_{X} = \frac{-\pi^{2} \beta^{2} K_{\rm N} L (\gamma - 1)(v_{\rm w1} + v_{\rm w2})}{K_{\rm N} L (\gamma + 2 \pi^2 \beta^{2}) + 3 \pi \mu a (1 + 2 \pi^2 \beta^{2})},
	\label{eqn:Ux}
\end{equation}	
where $v_{\rm{w}k} = \lambda f_{k}$ is the wave propagation velocity of the flagellum, $\gamma = K_{\rm{L}}/K_{\rm{N}}$ is the drag coefficient ratio, $\beta = A/\lambda$ is the flagellar shape coefficient.

We can derive $P_{1} = \int_{s_{1}=0}^{L}\vec{v}_{1} \cdot \mathrm{d}\vec{F}_{1}$ as the power produced by the anterior flagellum, where $\vec{v}_{1} = \vec{v}_{\rm w/f(1)}$ as the relative velocity vector of fluid to each flagellar segment, and $\mathrm{d}\vec{F}_{1}$ is the drag force of fluid acting on each flagellar segment. We can then rewrite $P_1$ as
\begin{equation}
	P_1 = K_{\rm N1} L_1 \left[ \left( \gamma_{1}-1 \right) \frac{\left( 2 \pi^2 v_{\rm w1} \beta_{1}^2 - U_{X} \right)^2}{1 + 2 \pi^2 \beta_{1}^2} + U_{X}^2 + 2 \pi^2 v_{\rm w1}^2 \beta_{1}^2 \right].
	\label{eqn:P1}
\end{equation}

Similarly, we can also derive power produced by the poterior flagellum $P_2$ as
\begin{equation}
	P_2 = K_{\rm N2} L_2 \left[ \left( \gamma_{2}-1 \right) \frac{\left( 2 \pi^2 v_{\rm w2} \beta_{2}^2 + U_{X} \right)^2}{1 + 2 \pi^2 \beta_{2}^2} + U_{X}^2 + 2 \pi^2 v_{\rm w2}^2 \beta_{2}^2 \right].
	\label{eqn:P2}
\end{equation}

The useful power $P_0$ required to propel the cell body with a speed $U_X$ can be derived as
\begin{equation}
	P_0 = F_{d,cell} \cdot U_X = 6 \pi \mu a U_X^2.
	\label{eqn:P0}
\end{equation}

The efficiency of the two flagella in propelling the cell body is estimated as
\begin{equation}
 	\eta = \frac{P_0}{P_1 + P_2}.
 	\label{eqn:eta}
\end{equation}

%% Add this line AFTER all your figures and tables
\FloatBarrier
% \pagebreak
\movie{(Zoospore Robot Design)} The movie illustrates the zoospore-inspired robot's biflagellated design, demonstrating its high-speed propulsion and maneuverability through coordinated flagellar motion in a viscous environment.
\\

\movie{(Experiments on Propulsion Velocity vs Flagellar Length)}
The movie showcases experiments investigating how varying flagellar length affects the propulsion velocity of the zoospore-inspired robot, demonstrating that longer flagella enhances propulsion speed in viscous environments.
\\

\movie{(Experiments on Propulsion Velocity vs Flagellar Beating Frequency)}
The movie illustrates experiments examining the impact of different flagellar beating frequencies on the propulsion velocity of the zoospore-inspired robot, highlighting the optimal frequency range for maximizing speed.
\\

\movie{(Experiments on Propulsion Velocity vs Flagellar Arrangement)}
The movie demonstrates experiments assessing how different flagellar arrangements influence the propulsion velocity of the zoospore-inspired robot, revealing the effects of anterior, posterior, and dual-flagella configurations on speed and efficiency.

%\dataset{dataset_one.txt}{Type or paste legend here.}

%\dataset{dataset_two.txt}{Type or paste legend here. Adding longer text to show what happens, to decide on alignment and/or indentations for multi-line or paragraph captions.}

%\pagebreak
\bibliography{zoospore_bib}